\documentclass{article}

\PassOptionsToPackage{numbers, compress}{natbib}
\usepackage[preprint]{neurips_2026}
\usepackage[utf8]{inputenc}
\usepackage[T1]{fontenc}
\usepackage{hyperref}
\usepackage{url}
\usepackage{booktabs}
\usepackage{amsfonts}
\usepackage{nicefrac}
\usepackage{microtype}
\usepackage{xcolor}
\usepackage{graphicx}
\usepackage{multirow}
\usepackage{tikz}
\usetikzlibrary{arrows.meta,positioning}

\title{GraphRAG on Consumer Hardware: Benchmarking Local LLMs for Healthcare EHR Schema Retrieval}

\author{%
  Peter Fernandes \\
  Department of Computer Engineering \\
  California Polytechnic State University \\
  San Luis Obispo, CA, USA \\
  \texttt{pfernand@calpoly.edu} \\
  \And
  Ria Kanjilal \\
  Department of Computer Engineering \\
  California Polytechnic State University \\
  San Luis Obispo, CA, USA \\
  \texttt{rkanjila@calpoly.edu} \\
}

\begin{document}

\maketitle

\begin{abstract}
Graph-based Retrieval Augmented Generation (GraphRAG) extends retrieval-augmented generation to support structured reasoning over complex corpora, but its reliability under resource-constrained, privacy-sensitive deployments remains unclear. In healthcare, where Electronic Health Record (EHR) data is complex and strictly regulated, reliance on cloud-based large language models (LLMs) introduces challenges in cost, latency, and compliance. In this work, we present a systematic evaluation of GraphRAG for EHR schema retrieval using locally deployed open-source LLMs. We implement the Microsoft GraphRAG pipeline on real-world EHR schema documentation and benchmark four models, including Llama 3.1 (8B), Mistral (7B), Qwen 2.5 (7B), and Phi-4-mini (3.8B), each deployed via Ollama on a single consumer GPU (8 GB VRAM). We evaluate indexing efficiency, knowledge graph construction, query latency, answer quality, and hallucination under both global and local retrieval modes. Our results reveal substantial differences: Llama 3.1 produces the richest knowledge graph (1,172 entities), Qwen 2.5 achieves the best answer quality (3.3/5), Phi-4-mini fails to complete the pipeline due to structured-output errors, and Mistral exhibits degenerate repetition behavior.  We further show that GraphRAG exhibits a practical capacity threshold, where models below approximately 7B parameters fail to reliably produce valid structured outputs and cannot complete the pipeline. In addition, indexing and answer quality are decoupled across models, and local retrieval consistently outperforms global summarization in both latency and factual grounding, with reduced hallucination. These findings demonstrate that GraphRAG is feasible on consumer hardware while highlighting the importance of model selection and retrieval design for robust deployment in regulated settings.
\end{abstract}

\section{Introduction}
\label{sec:intro}

Retrieval Augmented Generation (RAG)~\cite{lewis2020retrieval} grounds language model outputs in retrieved evidence, enabling question answering over large domain-specific corpora without full fine-tuning. Standard RAG encodes documents into dense vector embeddings and retrieves the top-$k$ most similar chunks at query time. While effective for simple factual lookups, this flat approach struggles with queries requiring multi-hop reasoning or a global understanding of entity relationships~\cite{edge2024local}. GraphRAG, introduced by Edge et al.~\cite{edge2024local}, addresses these limitations by constructing a knowledge graph during an offline indexing phase. In this process, entities and relationships are first extracted from text chunks and assembled into a graph representation, which is then organized into communities using the Leiden algorithm~\cite{traag2019louvain} and summarized hierarchically to capture structure at multiple levels of abstraction. At query time, \emph{local search} retrieves entity neighborhoods while \emph{global search} leverages community-level summaries for corpus-wide synthesis.

In many real-world domains, including enterprise data management and large-scale information systems, structured schema documentation is a critical yet complex knowledge resource, often comprising thousands of interrelated tables with intricate dependencies \cite{halevy2005enterprise, stonebraker2018end}. While traditional retrieval methods can access relevant fragments, they are limited in capturing relationships across entities or enabling global synthesis \cite{lewis2020retrieval,edge2024local,gao2023retrieval}. These challenges are particularly pronounced in healthcare informatics. Electronic health record (EHR) systems rely on large, highly structured schemas to organize clinical and operational data, where accurate interpretation of inter-table relationships is essential for analytics, reporting, and decision-support workflows \cite{hripcsak2015observational}. As a representative example, Epic's Clarity data model provides a large-scale relational schema widely used in practice. Prior work has shown that large language models (LLMs) can encode substantial clinical knowledge~\cite{singhal2023large}, motivating their integration into healthcare systems. However, deploying LLM-based retrieval in this setting introduces practical constraints. The GraphRAG indexing phase requires numerous LLM calls, making cloud-based deployment costly at scale \cite{achiam2023gpt,zhao2023survey}. Moreover, sending schema metadata to external providers raises compliance concerns under regulations such as Health Insurance Portability and Accountability Act (HIPAA), while reliance on remote application programming interfaces (API) introduces latency that can hinder interactive applications \cite{price2019privacy,patterson2021carbon, zhao2023survey}. Recent advances in efficient open-source models, deployable locally via Ollama, offer a promising alternative through on-premise inference on consumer hardware \cite{grattafiori2024llama,yang2025qwen3}. However, it remains unclear whether models at the 7B parameter scale can reliably support the structured extraction, graph construction, and multi-stage reasoning required by GraphRAG pipelines \cite{edge2024local}.

In this study, we present an empirical evaluation of GraphRAG applied to EHR schema retrieval using locally deployed LLMs. We implement the Microsoft GraphRAG pipeline on real-world Epic Clarity schema documentation and benchmark four open-source models such as Llama 3.1 (8B), Mistral (7B), Qwen 2.5 (7B), and Phi-4-mini (3.8B) on a single consumer GPU~\cite{jiang20236g,grattafiori2024llama,abdin2024phi,yang2025qwen3}. We evaluate performance across indexing efficiency, knowledge graph construction, query latency, answer quality, and hallucination behavior under both global and local retrieval modes.

Our contributions can be summarized as follows:
\begin{itemize}
    \item We present an empirical evaluation of GraphRAG applied to EHR schema documentation, using Epic Clarity as a representative real-world system. Unlike prior work focused on general-domain corpora or task-specific benchmarks, we analyze GraphRAG as a full pipeline in a structured healthcare setting under local deployment constraints, focusing on pipeline-level reliability rather than benchmarking alternative RAG variants.
    
    \item We conduct a controlled comparison of four locally deployed open-source LLMs across the GraphRAG pipeline on a single consumer GPU, evaluating indexing efficiency, graph construction, query latency, and answer quality under both local and global retrieval modes.
    
    \item We identify critical failure modes in resource-constrained settings, including structured-output failures in smaller models and degenerate repetition behavior, highlighting robustness limitations of local GraphRAG deployments.
    
    \item We demonstrate that indexing quality and query quality are \emph{decoupled}, showing that stronger entity extraction and larger graphs do not necessarily yield better answers, suggesting stage-specific model strengths within the pipeline.
    
    \item We analyze deployment trade-offs in cost, privacy, and system design, showing that local GraphRAG eliminates API costs and data egress while introducing model-dependent constraints, providing practical guidance for deployment in regulated environments.
\end{itemize}

The remainder of the paper is organized as follows: Section~\ref{sec:relatedwork} reviews related work; Section~\ref{sec:method} presents the methodology, followed by Section~\ref{sec:experimentalsetup}, which describes the experimental setup; Sections~\ref{sec:results} and~\ref{sec:discussion} present and discuss the results; and Section~\ref{sec:conclusion} concludes the paper.

\section{Related Work}
\label{sec:relatedwork}

Prior work on retrieval-augmented generation has extensively studied methods for improving grounding and retrieval quality in LLMs. Gao et al.~\cite{gao2023retrieval} categorize RAG systems into naive, advanced, and modular variants, and highlight limitations in handling queries that require multi-hop reasoning and cross-document synthesis. While these approaches improve retrieval effectiveness, they largely operate over unstructured text and do not explicitly model relationships between entities. Graph-based extensions to RAG address this limitation by incorporating structured representations. In~\cite{edge2024local}, the authors demonstrate that organizing retrieved information into a knowledge graph with community-level summaries can significantly improve performance on global sensemaking tasks. Subsequent benchmarking efforts~\cite{cai2025mollangbench,han2025rag} further show the advantages of graph-structured retrieval over standard RAG in complex query settings. However, these studies focus primarily on general-domain corpora and cloud-based LLMs, and do not examine the reliability of GraphRAG pipelines under constrained computational settings or domain-specific structured data such as database schemas. A parallel line of work explores the integration of LLMs with knowledge graphs. Pan et al.~\cite{pan2024unifying} outline multiple paradigms for combining symbolic and neural representations, including LLM-augmented knowledge graphs where models are used to construct structured representations from text. Existing work in this space primarily emphasizes improvements in reasoning or knowledge integration, whereas our work focuses on system-level behavior, particularly the robustness of multi-stage pipelines involving structured extraction and retrieval.

Recent advances in efficient open-source LLMs have made it feasible to run strong models on commodity hardware, enabling local inference in resource-constrained settings. Models such as Llama~3.1~\cite{grattafiori2024llama}, Mistral~7B~\cite{jiang20236g}, Phi-4~\cite{abdin2024phi}, and Qwen~2.5~\cite{yang2025qwen3} demonstrate competitive performance at relatively small scales. However, existing evaluations are largely limited to standard benchmarks and do not consider structured, multi-stage pipelines such as GraphRAG, where reliability depends on both generation quality and the ability to produce valid structured outputs for downstream graph construction. While prior work has advanced both retrieval-augmented generation and efficient LLM deployment, the behavior of GraphRAG under local deployment constraints, particularly for structured schema reasoning in domains such as healthcare, remains underexplored.

\section{Methodology}
\label{sec:method}

\subsection{Dataset}

Our corpus consists of HTML files exported from Epic's DocGen tool, each documenting a single Clarity database table, including its name and description, column definitions (name, data type, constraints), foreign-key relationships, and join conditions. The full dataset contains over 7{,}000 files. For this study, we use a curated 10-file subset covering account management and clinical notification tables: \texttt{ABN\_FOLLOW\_UP}, \texttt{ABN\_MEDICATIONS}, \texttt{ABN\_NOTES}, \texttt{ABN\_ORDERS}, \texttt{ABN\_ORDER\_INFO}, \texttt{ACCESSIBLE\_SERVICES}, \texttt{ACCOUNT}, \texttt{ACCOUNT\_2}, \texttt{ACCOUNT\_3}, and \texttt{ACCOUNT\_CONTACT}. After HTML parsing, these yield eight documents comprising 141 text units. Although this subset is small, it preserves key structural characteristics of EHR schemas, including inter-table dependencies and relational patterns, and enables controlled analysis of pipeline-level behavior.

\subsection{GraphRAG Pipeline}

\begin{figure}[!t]
    \centering
    \includegraphics[width=0.8\linewidth]{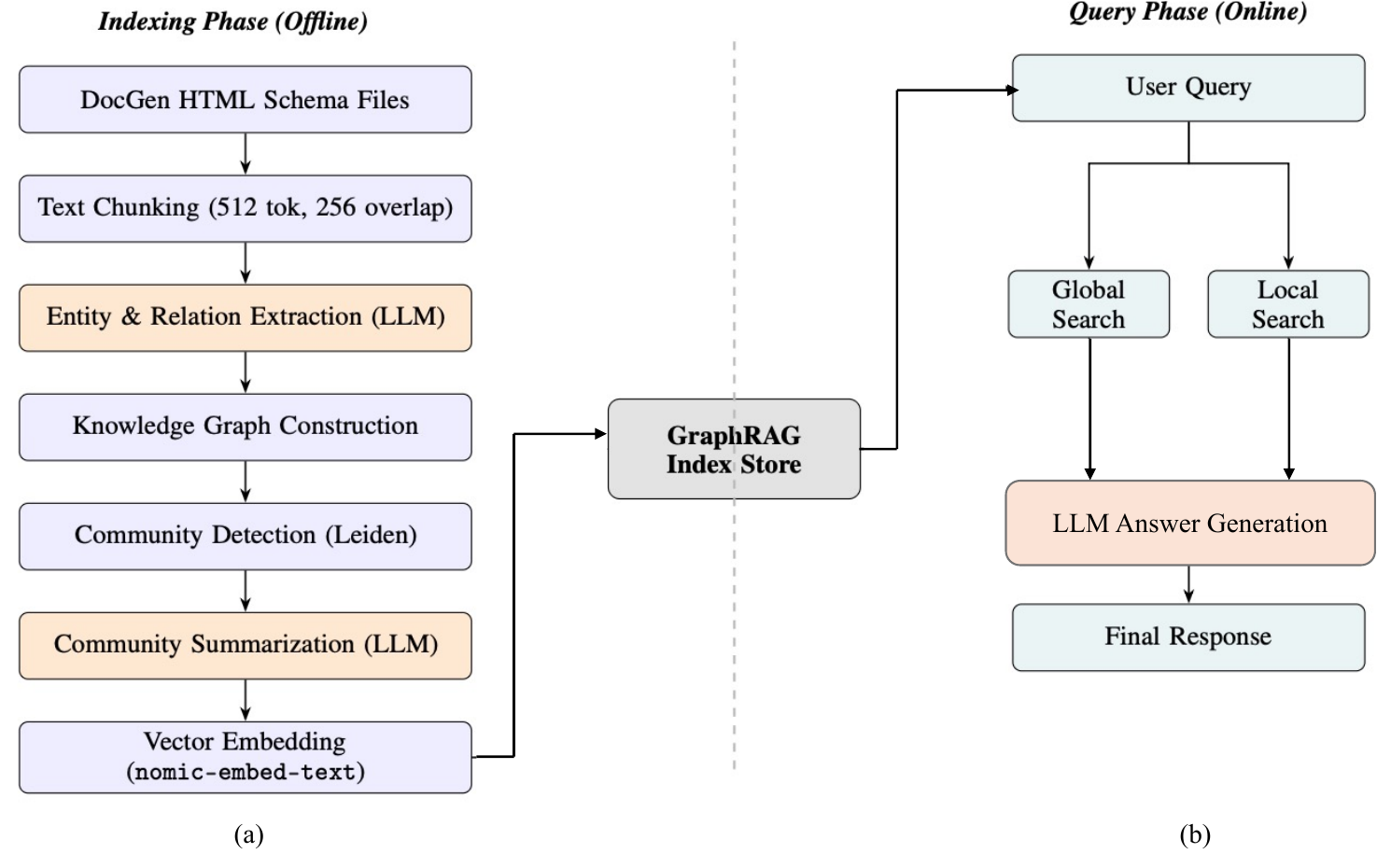}
    \caption{GraphRAG system architecture. (a) Offline indexing: HTML schema files are chunked and processed by a language model to extract entities and relationships, which are assembled into a knowledge graph, organized into communities using the Leiden algorithm, summarized hierarchically, and embedded into an index store. (b) Online querying: a user query is processed through either global search, which aggregates community-level summaries, or local search, which retrieves entity neighborhoods, followed by LLM-based answer generation.}
\label{fig:system}
\end{figure}

We use Microsoft GraphRAG~v2.3.0~\cite{edge2024local} and implement a six-stage pipeline. First, the input HTML schema files are segmented into 512-token chunks with 256-token overlap to preserve context continuity. Next, an LLM extracts entities and relationships, including tables, columns, data types, and foreign-key links, from each chunk. The extracted information is then used to construct a knowledge graph composed of structured triples. This graph is subsequently organized into hierarchical communities using the Leiden algorithm~\cite{traag2019louvain}. At each level, the model generates natural language summaries to capture community-level semantics. Finally, both text units and entity descriptions are embedded using \texttt{nomic-embed-text}~\cite{nussbaum2024nomic} (274\,MB) to support efficient vector-based retrieval during querying. Figure~\ref{fig:system} illustrates the overall architecture, highlighting the separation between the offline indexing phase and the online query phase.

\subsection{Models Under Evaluation}

We evaluate four open-source LLMs that reflect different design priorities in instruction tuning, reasoning, and efficient deployment. Llama~3.1~\cite{grattafiori2024llama} and Qwen~2.5~\cite{yang2025qwen3} are instruction-tuned models with strong alignment and reasoning capabilities. Mistral~\cite{jiang20236g} emphasizes architectural efficiency for high performance with reduced compute, while Phi-4-mini~\cite{abdin2024phi} is designed for lightweight deployment in resource-constrained environments. Together, these models provide a representative set for evaluating GraphRAG under local deployment settings. Table~\ref{tab:models} summarizes the models, all deployed using Ollama~v0.17.0 with Q4\_K\_M quantization. The same embedding model, \texttt{nomic-embed-text}~\cite{nussbaum2024nomic}, is used consistently across all pipelines.

\begin{table}[h]
  \caption{Models evaluated in this study.}
  \label{tab:models}
  \centering
  \begin{tabular}{lrrr}
    \toprule
    Model        & Parameters & Context & Disk size \\
    \midrule
    Llama 3.1~\cite{grattafiori2024llama}  & 8B         & 128K    & 4.9\,GB  \\
    Mistral~\cite{jiang20236g}   & 7B         & 32K     & 4.4\,GB  \\
    Phi-4-mini~\cite{abdin2024phi}& 3.8B       & 128K    & 2.5\,GB  \\
    Qwen 2.5~\cite{yang2025qwen3}  & 7B         & 128K    & 4.7\,GB  \\
    \bottomrule
  \end{tabular}
\end{table}

\subsection{Evaluation Queries}

We evaluate three representative queries designed to test different retrieval capabilities:
\begin{itemize}
  \item \textbf{Q1 (broad discovery):} ``What tables store patient demographic information?''
  \item \textbf{Q2 (relationship understanding):} ``Describe the relationships between
        account-related tables.''
  \item \textbf{Q3 (specific detail):} ``What columns are used to link ABN tables together?''
\end{itemize}
Each query is issued in both \emph{global search} (community-level map-reduce summarization) and \emph{local search} (entity-neighborhood vector retrieval) modes. These queries are designed to evaluate distinct retrieval behaviors, including broad schema discovery, relationship reasoning, and fine-grained linkage within the pipeline.

\subsection{Metrics}

We evaluate performance using several metrics. Indexing time measures the wall-clock duration required to complete the full GraphRAG indexing pipeline. Graph statistics capture the number of entities, relationships, and communities extracted during graph construction. Query latency is defined as the time from query submission to the first complete response. Answer quality is assessed manually by the authors based on correctness, completeness, and relevance, using a 1--5 scoring scale. Scores are assigned using consistent criteria of correctness, completeness, and relevance, and are applied uniformly across all models and queries. Finally, hallucination detection examines whether generated responses reference entities that are absent from the source files~\cite{ji2023survey}.

\section{Experimental Setup}
\label{sec:experimentalsetup}

\paragraph{Hardware.}
All experiments are conducted on a single consumer workstation equipped with an Intel Core i7-10700KF (8 cores / 16 threads, 3.8\,GHz), 32\,GB DDR4 memory, an NVIDIA GeForce GTX~1070~Ti (8\,GB VRAM), NVMe SSD storage, and running Windows~11 Home.

\paragraph{Software.}
The software environment consists of Python~3.12, Microsoft GraphRAG~v2.3.0, and Ollama~v0.17.0.

\paragraph{Pipeline configuration.}
Due to single-GPU memory constraints, concurrency is set to 1. Each model uses a separate output directory to prevent index contamination. The pipeline configuration is shown below:

{\small
\begin{verbatim}
models:
  default_chat_model:
    type: openai_chat
    api_base: http://localhost:11434/v1
    model: <model_name>
    model_supports_json: true
    concurrent_requests: 1
  default_embedding_model:
    type: openai_embedding
    api_base: http://localhost:11434/v1
    model: nomic-embed-text
\end{verbatim}
}

\section{Results}
\label{sec:results}

\subsection{Indexing Performance}

Table~\ref{tab:indexing} reports indexing results for all four models. Phi-4-mini crashed during community report generation with a \texttt{FailedToGenerateValidJsonError} after extracting only 16 entities and is therefore excluded from query evaluation. Llama~3.1 extracted 1{,}172 entities, nearly $2\times$ Mistral's 649 and $3.5\times$ Qwen~2.5's 330, at the cost of $2.4\times$ longer indexing time. Qwen~2.5 completed indexing in 88 minutes, the fastest among all models. Mistral identifies the highest number of relationships (926) despite extracting fewer entities, suggesting a more relation-focused rather than entity-centric extraction behavior.

\begin{table}[h]
  \caption{Indexing performance on the 10-file subset (8 documents, 141 text units). Phi-4-mini failed and is excluded from query evaluation.}
  \label{tab:indexing}
  \centering
  \begin{tabular}{lrrrrl}
    \toprule
    Model             & Time (min)      & Entities         & Relations        & Communities      & Status          \\
    \midrule
    Llama 3.1 (8B)    & 211.5           & \textbf{1{,}172} & 696              & 113              & Complete        \\
    Mistral (7B)      & 204             & 649              & \textbf{926}     & \textbf{123}     & Complete        \\
    Phi-4-mini (3.8B) & $\approx$12     & 16               & 5                & 1                & \textbf{Failed} \\
    Qwen 2.5 (7B)     & \textbf{88}     & 330              & 333              & 30               & Complete        \\
    \bottomrule
  \end{tabular}
\end{table}

\subsection{Query Latency}

Table~\ref{tab:latency} shows per-query latency. Local search remains consistently fast (30--34\,s) across all models. In contrast, global search exhibits significant variation: Mistral averages 281\,s due to processing 123 community reports, while Llama~3.1 and Qwen~2.5 average 110\,s and 156\,s, respectively. The Mistral Q2 local outlier (1{,}293\,s) corresponds to the degenerate repetition event described in Section~\ref{sec:failure}.

\begin{table}[h]
  \caption{Query latency in seconds. Mistral Q2 local (${}^{*}$) produced a degenerate repetition loop running for 21 minutes; excluded from Mistral local averages.}
  \label{tab:latency}
  \centering
  \small
  \begin{tabular}{lrrrrrr}
    \toprule
    & \multicolumn{2}{c}{Llama 3.1} & \multicolumn{2}{c}{Mistral} & \multicolumn{2}{c}{Qwen 2.5} \\
    \cmidrule(lr){2-3}\cmidrule(lr){4-5}\cmidrule(lr){6-7}
    Query & Global & Local & Global & Local            & Global & Local \\
    \midrule
    Q1    & 93     & 32    & 295    & 31               & 167    & 32    \\
    Q2    & 129    & 34    & 279    & 1{,}293$^{*}$    & 153    & 94    \\
    Q3    & 108    & 31    & 269    & 30               & 148    & 31    \\
    \midrule
    Avg   & 110    & 32    & 281    & $\approx$31      & 156    & 52    \\
    \bottomrule
  \end{tabular}
\end{table}

\subsection{Answer Quality}

Table~\ref{tab:quality} presents manual quality scores (1=poor, 5=excellent). Qwen~2.5 achieves the highest overall performance, with scores of 3.0/5 for global search and 3.3/5 for local search, despite extracting the fewest entities during indexing. This \emph{decoupling} between indexing quality and query quality is a central observation of this study. Llama~3.1 attains scores of 2.3 (global) and 2.7 (local), while Mistral performs worst, with a global average of 2.0.

\begin{table}[h]
  \caption{Manual answer quality (1=poor, 5=excellent) scored on correctness, completeness, and relevance. (${}^{*}$) Mistral Q2 local was a degenerate repetition output.}
  \label{tab:quality}
  \centering
  \small
  \begin{tabular}{lrrrrrr}
    \toprule
    & \multicolumn{2}{c}{Llama 3.1} & \multicolumn{2}{c}{Mistral} & \multicolumn{2}{c}{Qwen 2.5} \\
    \cmidrule(lr){2-3}\cmidrule(lr){4-5}\cmidrule(lr){6-7}
    Query & Global & Local & Global & Local      & Global       & Local        \\
    \midrule
    Q1    & 2      & 2     & 2      & 1          & \textbf{3}   & 2            \\
    Q2    & 2      & 3     & 2      & 1$^{*}$    & \textbf{3}   & \textbf{4}   \\
    Q3    & 3      & 3     & 2      & ---        & \textbf{3}   & \textbf{4}   \\
    \midrule
    Avg   & 2.3    & 2.7   & 2.0    & 1.0        & \textbf{3.0} & \textbf{3.3} \\
    \bottomrule
  \end{tabular}
\end{table}

\subsection{Hallucination Analysis}

During global search, all three successful models produce hallucinated database table names that are absent from the test corpus, consistent with known patterns in hallucination behavior~\cite{ji2023survey}. Llama~3.1 generates entities such as \texttt{PATIENT}, \texttt{PERSON}, \texttt{TRANSACTIONS}, \texttt{BALANCES}, and \texttt{CUSTOMERS}, while Mistral hallucinates \texttt{PATIENT\_DEMOGRAPHICS}, \texttt{ADDRESS}, \texttt{INSURANCE}, and \texttt{PAYMENT\_HISTORY}. Qwen~2.5 partially hallucinates \texttt{claim\_id}. In contrast, local search results remain grounded, with all models referencing actual schema entities such as \texttt{ABN\_FOLLOW\_UP}, \texttt{NOTE\_CSN\_ID}, and \texttt{PAT\_ID}. This pattern suggests that global search can cause 7B models to revert to training-data priors about ``typical'' databases rather than faithfully representing indexed content.

\subsection{Failure Modes}
\label{sec:failure}

\paragraph{Pipeline failure (Phi-4-mini, 3.8B).}
At 3.8B parameters (Q4\_K\_M quantized), the model could not reliably produce the structured JSON output required during entity extraction, yielding only 16 entities and 5 relationships. The pipeline subsequently crashed at the community report generation stage with \texttt{FailedToGenerateValidJsonError}. This suggests a practical minimum model size of approximately 7B parameters for reliable GraphRAG operation.

\paragraph{Degenerate repetition (Mistral, Q2 local search).}
For the account-relationships query under local search, Mistral entered an infinite repetition loop, generating increasingly incoherent column names
(``Next Payment Dispute Chargeback Reversal Reversal Reversal$\ldots$'') for over 21 minutes. This behavior aligns with known autoregressive failure modes~\cite{ji2023survey}, where repetitive context induces repetitive generation; it was exacerbated by Ollama's default configuration
lacking a repetition penalty.

\subsection{Cost Comparison}

Table~\ref{tab:cost} compares projected costs for local and cloud-based deployment. The local pipeline eliminates per-call API costs and avoids data egress to external (cloud) providers, which is particularly advantageous under HIPAA constraints.

\begin{table}[h]
  \caption{Cost comparison: local (Ollama) vs.\ projected cloud (OpenAI API) for a 10-file indexing run and ongoing query workload.}
  \label{tab:cost}
  \centering
  \begin{tabular}{lrr}
    \toprule
    Metric                       & Local (Ollama)        & Cloud (projected)  \\
    \midrule
    Indexing cost (10 files)     & \$0.00                & \$0.50--2.00       \\
    Query cost (per query)       & \$0.00                & \$0.01--0.05       \\
    Monthly est.\ (100 q/day)    & \$5--10 (electricity) & \$30--150          \\
    \bottomrule
  \end{tabular}
\end{table}

\section{Discussion}
\label{sec:discussion}

The results reveal a clear separation between indexing behavior and downstream answer quality. Qwen~2.5 extracts the fewest entities (330) yet produces the best answers (3.3/5 on local search), whereas Llama~3.1 extracts $3.5\times$ more entities but scores only 2.7/5. This indicates that a model's ability to follow extraction prompts and emit valid JSON does not necessarily predict its effectiveness in answer generation. A practical implication is that a hybrid pipeline, in which indexing is performed with Llama~3.1 for coverage and querying is handled by Qwen~2.5 for answer quality, may outperform any single-model configuration.

A similar pattern is observed when comparing retrieval modes. Qwen~2.5 achieves higher answer quality with local search (3.3/5) than with global search (3.0/5), while also delivering approximately $3\times$ lower latency ($\approx$52\,s vs.\ 156\,s) and substantially fewer hallucinations. Although GraphRAG is designed to excel at global sensemaking~\cite{edge2024local}, these results suggest that models at the 7B scale lack the capacity to faithfully synthesize information from dozens of community summaries. In contrast, retrieval over entity neighborhoods provides a more reliable approach on consumer hardware.

Model limitations further reinforce this behavior. The failure of Phi-4-mini establishes that models below approximately 7B parameters (Q4 quantized) cannot reliably complete the GraphRAG pipeline, highlighting a practical constraint for edge and resource-constrained deployments.

From a deployment perspective, the local pipeline offers additional advantages. Keeping all data on-premise eliminates the need for Business Associate Agreements (BAAs) with cloud LLM providers under HIPAA. For healthcare organizations handling schema metadata~\cite{singhal2023large}, this may constitute a compliance requirement that mandates local deployment regardless of quality tradeoffs. These cost advantages should be interpreted alongside observed performance trade-offs across models and retrieval strategies.

Taken together, these findings highlight several practical considerations for deployment. Qwen~2.5 consistently delivers the strongest answer quality on consumer hardware, whereas Llama~3.1 is more suitable when extraction coverage is the priority. Across all models, local search proves more reliable than global search, offering advantages in latency, grounding, and stability for local LLM deployments. The observed failure of Phi-4-mini further indicates that models below 7B parameters are not suitable for reliable GraphRAG operation in this setting. In addition, the degenerate repetition observed with Mistral indicates that inference configuration plays an important role, and enabling a repetition penalty in Ollama can help mitigate such unstable generation behavior.

\section{Conclusion}
\label{sec:conclusion}

We present a systematic evaluation of GraphRAG~\cite{edge2024local} for healthcare EHR schema retrieval under local deployment constraints. By implementing the full GraphRAG pipeline on real-world schema documentation and benchmarking four open-source LLMs on a single consumer GPU, this work provides a system-level characterization of model behavior across indexing, graph construction, and query stages. Our study highlights several key observations: (i) GraphRAG can be feasibly deployed on consumer hardware with models at the 7B scale and above, (ii) model performance varies substantially across pipeline stages, with Qwen~2.5 achieving the highest answer quality and Llama~3.1 producing the most extensive knowledge graph, (iii) models below approximately 3.8B parameters fail to reliably execute structured extraction, (iv) local retrieval consistently yields better latency and grounding than global summarization, and (v) hallucination remains a persistent issue in global search due to reliance on training-data priors. In addition, local deployment eliminates per-call API costs and avoids data egress, providing practical advantages in cost and compliance for regulated environments.

Beyond these findings, the work contributes a practical evaluation of GraphRAG in a structured, domain-specific setting where reliability, latency, and compliance constraints are critical. In particular, the work emphasizes pipeline-level behavior rather than isolated model performance. The results demonstrate that effective deployment depends not only on model capability but also on how different stages of the pipeline are configured, motivating the need for stage-aware design choices.

This study has several limitations. First, the evaluation is conducted on a curated 10-file subset of a larger schema corpus, which, while representative, does not capture the full scale and diversity of real-world deployments. Second, answer quality is assessed through manual scoring, which may introduce subjectivity despite consistent criteria. Third, the experiments are performed on a single hardware configuration, and performance may vary across different systems or scaling conditions.

Future work will extend this analysis to the full 7{,}000-file corpus and investigate hybrid pipeline designs that decouple indexing and querying across models. Additional directions include conducting formal human evaluation with clinical informatics professionals and establishing a cloud-based baseline for direct comparison of quality, latency, and cost trade-offs between local and API-based deployments.



{\small
\bibliographystyle{unsrtnat}
\bibliography{reference.bib}

@article{lewis2020retrieval,
  title={Retrieval-augmented generation for knowledge-intensive nlp tasks},
  author={Lewis, Patrick and Perez, Ethan and Piktus, Aleksandra and Petroni, Fabio and Karpukhin, Vladimir and Goyal, Naman and K{\"u}ttler, Heinrich and Lewis, Mike and Yih, Wen-tau and Rockt{\"a}schel, Tim and others},
  journal={Advances in neural information processing systems},
  volume={33},
  pages={9459--9474},
  year={2020}
}

@article{edge2024local,
  title={From local to global: A graph rag approach to query-focused summarization},
  author={Edge, Darren and Trinh, Ha and Cheng, Newman and Bradley, Joshua and Chao, Alex and Mody, Apurva and Truitt, Steven and Metropolitansky, Dasha and Ness, Robert Osazuwa and Larson, Jonathan},
  journal={arXiv preprint arXiv:2404.16130},
  year={2024}
}

@article{gao2023retrieval,
  title={Retrieval-augmented generation for large language models: A survey},
  author={Gao, Yunfan and Xiong, Yun and Gao, Xinyu and Jia, Kangxiang and Pan, Jinliu and Bi, Yuxi and Dai, Yixin and Sun, Jiawei and Wang, Haofen and Wang, Haofen and others},
  journal={arXiv preprint arXiv:2312.10997},
  volume={2},
  number={1},
  pages={32},
  year={2023}
}

@article{han2025rag,
  title={Rag vs. graphrag: A systematic evaluation and key insights},
  author={Han, Haoyu and Ma, Li and Wang, Yu and Shomer, Harry and Lei, Yongjia and Qi, Zhisheng and Guo, Kai and Hua, Zhigang and Long, Bo and Liu, Hui and others},
  journal={arXiv preprint arXiv:2502.11371},
  year={2025}
}

@article{pan2024unifying,
  title={Unifying large language models and knowledge graphs: A roadmap},
  author={Pan, Shirui and Luo, Linhao and Wang, Yufei and Chen, Chen and Wang, Jiapu and Wu, Xindong},
  journal={IEEE Transactions on Knowledge and Data Engineering},
  volume={36},
  number={7},
  pages={3580--3599},
  year={2024},
  publisher={IEEE}
}

@article{traag2019louvain,
  title={From Louvain to Leiden: guaranteeing well-connected communities},
  author={Traag, Vincent A and Waltman, Ludo and Van Eck, Nees Jan},
  journal={Scientific reports},
  volume={9},
  number={1},
  pages={5233},
  year={2019},
  publisher={Nature Publishing Group UK London}
}

@article{ji2023survey,
  title={Survey of hallucination in natural language generation},
  author={Ji, Ziwei and Lee, Nayeon and Frieske, Rita and Yu, Tiezheng and Su, Dan and Xu, Yan and Ishii, Etsuko and Bang, Ye Jin and Madotto, Andrea and Fung, Pascale},
  journal={ACM computing surveys},
  volume={55},
  number={12},
  pages={1--38},
  year={2023},
  publisher={ACM New York, NY}
}

@article{zhao2023survey,
  title={A survey of large language models},
  author={Zhao, Wayne Xin and Zhou, Kun and Li, Junyi and Tang, Tianyi and Wang, Xiaolei and Hou, Yupeng and Min, Yingqian and Zhang, Beichen and Zhang, Junjie and Dong, Zican and others},
  journal={arXiv preprint arXiv:2303.18223},
  volume={1},
  number={2},
  pages={1--124},
  year={2023}
}

@article{grattafiori2024llama,
  title={The llama 3 herd of models},
  author={Grattafiori, Aaron and Dubey, Abhimanyu and Jauhri, Abhinav and Pandey, Abhinav and Kadian, Abhishek and Al-Dahle, Ahmad and Letman, Aiesha and Mathur, Akhil and Schelten, Alan and Vaughan, Alex and others},
  journal={arXiv preprint arXiv:2407.21783},
  year={2024}
}

@article{jiang20236g,
  title={6G non-terrestrial networks enabled low-altitude economy: Opportunities and challenges},
  author={Jiang, Yihang and Li, Xiaoyang and Zhu, Guangxu and Li, Hang and Deng, Jing and Han, Kaifeng and Shen, Chao and Shi, Qingjiang and Zhang, Rui},
  journal={arXiv preprint arXiv:2311.09047},
  year={2023}
}

@article{abdin2024phi,
  title={Phi-4 technical report},
  author={Abdin, Marah and Aneja, Jyoti and Behl, Harkirat and Bubeck, S{\'e}bastien and Eldan, Ronen and Gunasekar, Suriya and Harrison, Michael and Hewett, Russell J and Javaheripi, Mojan and Kauffmann, Piero and others},
  journal={arXiv preprint arXiv:2412.08905},
  year={2024}
}

@article{yang2025qwen3,
  title={Qwen3 technical report},
  author={Yang, An and Li, Anfeng and Yang, Baosong and Zhang, Beichen and Hui, Binyuan and Zheng, Bo and Yu, Bowen and Gao, Chang and Huang, Chengen and Lv, Chenxu and others},
  journal={arXiv preprint arXiv:2505.09388},
  year={2025}
}

@article{nussbaum2024nomic,
  title={Nomic embed: Training a reproducible long context text embedder},
  author={Nussbaum, Zach and Morris, John X and Duderstadt, Brandon and Mulyar, Andriy},
  journal={arXiv preprint arXiv:2402.01613},
  year={2024}
}

@article{cai2025mollangbench,
  title={Mollangbench: A comprehensive benchmark for language-prompted molecular structure recognition, editing, and generation},
  author={Cai, Feiyang and Bai, Jiahui and Tang, Tao and He, Guijuan and Luo, Joshua and Zhu, Tianyu and Pilla, Srikanth and Li, Gang and Liu, Ling and Luo, Feng},
  journal={arXiv preprint arXiv:2505.15054},
  year={2025}
}

@article{singhal2023large,
  title={Large language models encode clinical knowledge},
  author={Singhal, Karan and Azizi, Shekoofeh and Tu, Tao and Mahdavi, S Sara and Wei, Jason and Chung, Hyung Won and Scales, Nathan and Tanwani, Ajay and Cole-Lewis, Heather and Pfohl, Stephen and others},
  journal={Nature},
  volume={620},
  number={7972},
  pages={172--180},
  year={2023},
  publisher={Nature Publishing Group UK London}
}

@inproceedings{halevy2005enterprise,
  title={Enterprise information integration: successes, challenges and controversies},
  author={Halevy, Alon Y and Ashish, Naveen and Bitton, Dina and Carey, Michael and Draper, Denise and Pollock, Jeff and Rosenthal, Arnon and Sikka, Vishal},
  booktitle={Proceedings of the 2005 ACM SIGMOD international conference on Management of data},
  pages={778--787},
  year={2005}
}

@incollection{stonebraker2018end,
  title={The end of an architectural era: it's time for a complete rewrite},
  author={Stonebraker, Michael and Madden, Samuel and Abadi, Daniel J and Harizopoulos, Stavros and Hachem, Nabil and Helland, Pat},
  booktitle={Making Databases Work: the Pragmatic Wisdom of Michael Stonebraker},
  pages={463--489},
  year={2018}
}

@article{hripcsak2015observational,
  title={Observational Health Data Sciences and Informatics (OHDSI): opportunities for observational researchers},
  author={Hripcsak, George and Duke, Jon D and Shah, Nigam H and Reich, Christian G and Huser, Vojtech and Schuemie, Martijn J and Suchard, Marc A and Park, Rae Woong and Wong, Ian Chi Kei and Rijnbeek, Peter R and others},
  journal={Studies in health technology and informatics},
  volume={216},
  pages={574},
  year={2015}
}

@article{achiam2023gpt,
  title={Gpt-4 technical report},
  author={Achiam, Josh and Adler, Steven and Agarwal, Sandhini and Ahmad, Lama and Akkaya, Ilge and Aleman, Florencia Leoni and Almeida, Diogo and Altenschmidt, Janko and Altman, Sam and Anadkat, Shyamal and others},
  journal={arXiv preprint arXiv:2303.08774},
  year={2023}
}

@article{price2019privacy,
  title={Privacy in the age of medical big data},
  author={Price, W Nicholson and Cohen, I Glenn},
  journal={Nature medicine},
  volume={25},
  number={1},
  pages={37--43},
  year={2019},
  publisher={Nature Publishing Group US New York}
}

@article{patterson2021carbon,
  title={Carbon emissions and large neural network training},
  author={Patterson, David and Gonzalez, Joseph and Le, Quoc and Liang, Chen and Munguia, Lluis-Miquel and Rothchild, Daniel and So, David and Texier, Maud and Dean, Jeff},
  journal={arXiv preprint arXiv:2104.10350},
  year={2021}
}
}

\appendix
\section{Pipeline Configuration Details}

The GraphRAG configuration used across all experiments sets chunk size to 512 tokens with
256-token overlap and \texttt{concurrent\_requests: 1} (single-GPU constraint). Each model
run used a separate output directory (\texttt{ollama-test-cache-<model>/}) to prevent index
contamination. Full configuration YAML files and per-query response logs are available upon
request.


\end{document}